\documentclass{article}
\usepackage{amsmath,epsfig}
\usepackage{amsfonts}
\usepackage{makecell}
\usepackage{subfigure}

\usepackage[preprint]{spconfa4}

\let\OLDthebibliography\thebibliography
\renewcommand\thebibliography[1]{
  \OLDthebibliography{#1}
  \setlength{\parskip}{0pt}
  \setlength{\itemsep}{0pt plus 0.3ex}
}

\begin{document}\sloppy
\topmargin=0mm

\def\x{{\mathbf x}}
\def\L{{\cal L}}

\title{Multi-Task Learning via Co-Attentive Sharing for 
Pedestrian Attribute Recognition}
%
\name{Haitian Zeng, Haizhou Ai, Zijie Zhuang, Long Chen}

\address{Beijing National Research Center for Information Science and Technology, \\ Department of Computer Science and Technology, Tsinghua University, Beijing, China \\  \{zht17, zhuangzj15, l-chen16\}@mails.tsinghua.edu.cn, ahz@mail.tsinghua.edu.cn}

\maketitle

\begin{abstract}
Learning to predict multiple attributes of a pedestrian is a 
multi-task learning problem. 
To share feature representation between two individual task networks,
conventional methods like Cross-Stitch \cite{misra2016cross} and Sluice \cite{ruder122017sluice} network 
learn a linear combination of features or feature subspaces. 
However, linear combination rules out the complex interdependency
between channels.
Moreover, spatial information exchanging is less-considered.
In this paper, we propose a novel Co-Attentive Sharing (CAS) module 
which extracts discriminative channels and spatial regions for more
effective feature sharing in multi-task learning.
The module consists of three branches, which leverage different 
channels for between-task feature fusing, attention generation and 
task-specific feature enhancing, respectively.
Experiments on two pedestrian attribute recognition datasets show
that our module outperforms the conventional sharing units
and achieves superior results compared to the state-of-the-art 
approaches using many metrics.
\end{abstract}

\section{Introduction}
\label{sec:intro}
Recognizing person attributes has attracted great interest recently.
Given an image of a single person,
the aim is to recognize a series of semantic attributes, \emph{e.g.} age,
gender, clothing style, \emph{etc}.
It has a wide range of application scenarios since it creates a rich 
profile of human traits which can facilitate 
person retrieval \cite{sun2017svdnet} or re-identification \cite{lin2019improving}.

\begin{figure}
    \centering
    \subfigure[Hard-Sharing structure]
    {
        \includegraphics[width=4cm]{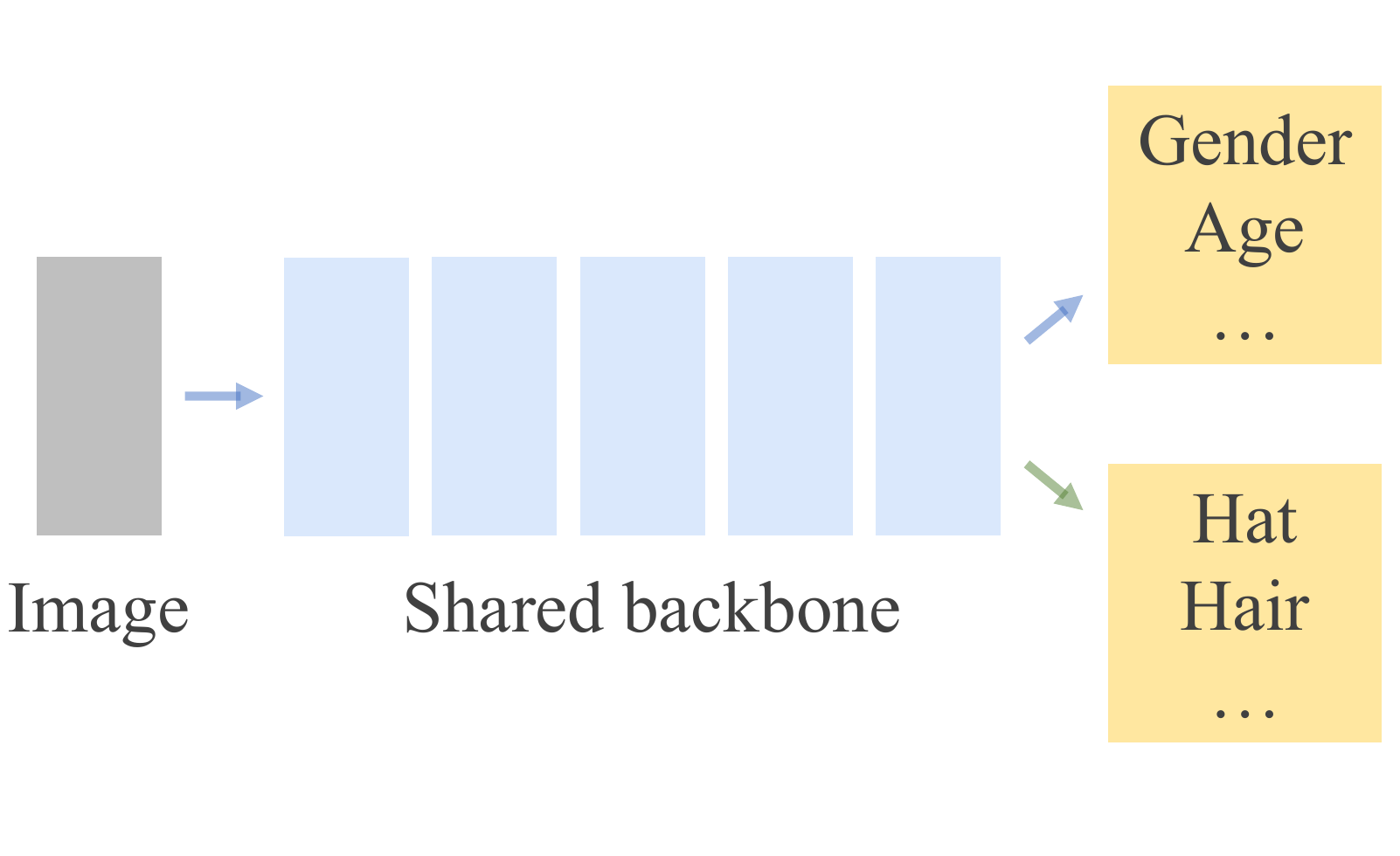}
        \label{fig:shared_structure}
    }
    \subfigure[Vanilla structure]
    {
        \includegraphics[width=4cm]{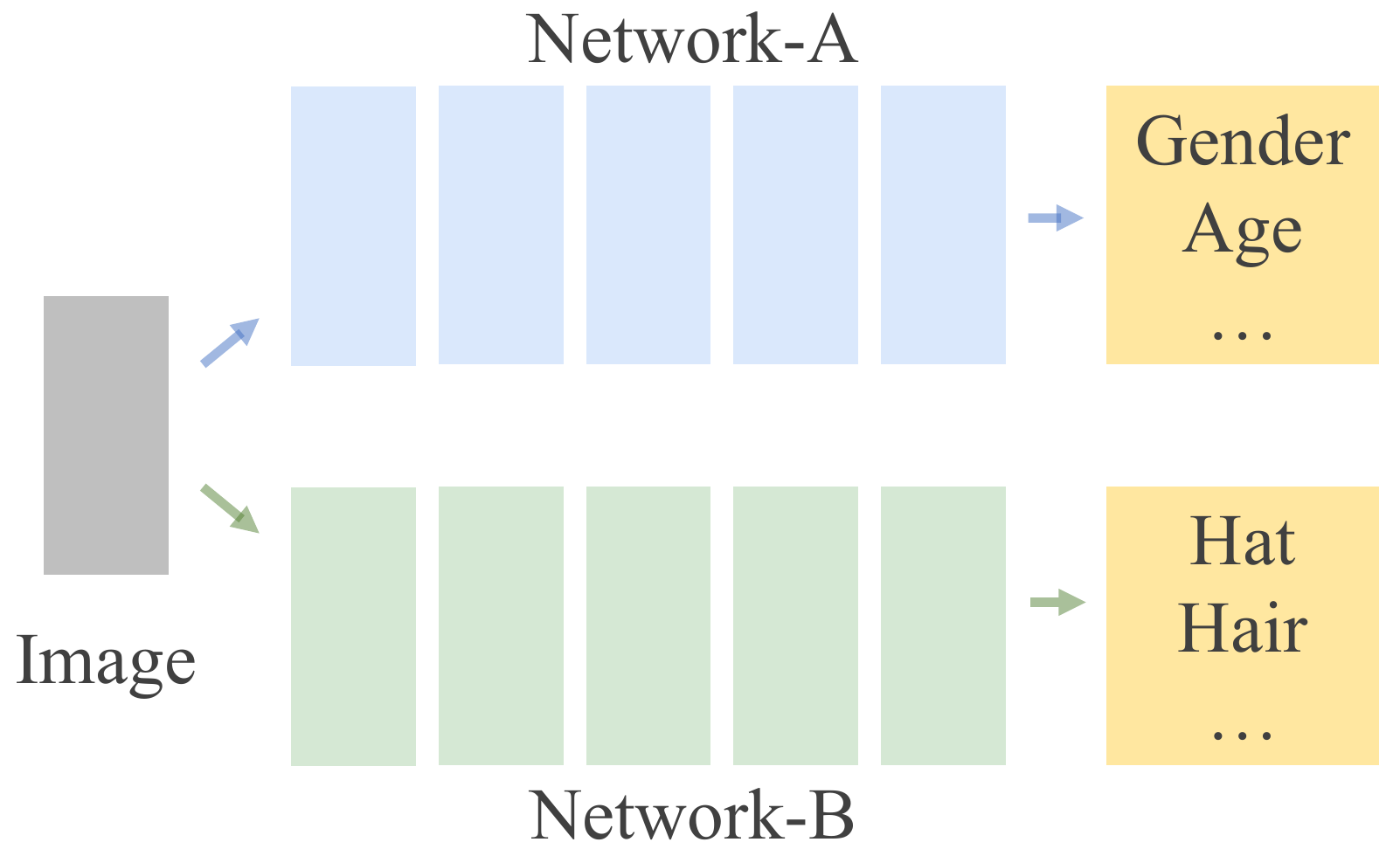}
        \label{fig:vanilla_structure}
    }
    \subfigure[Soft-Sharing structure]
    {
        \includegraphics[width=8.5cm]{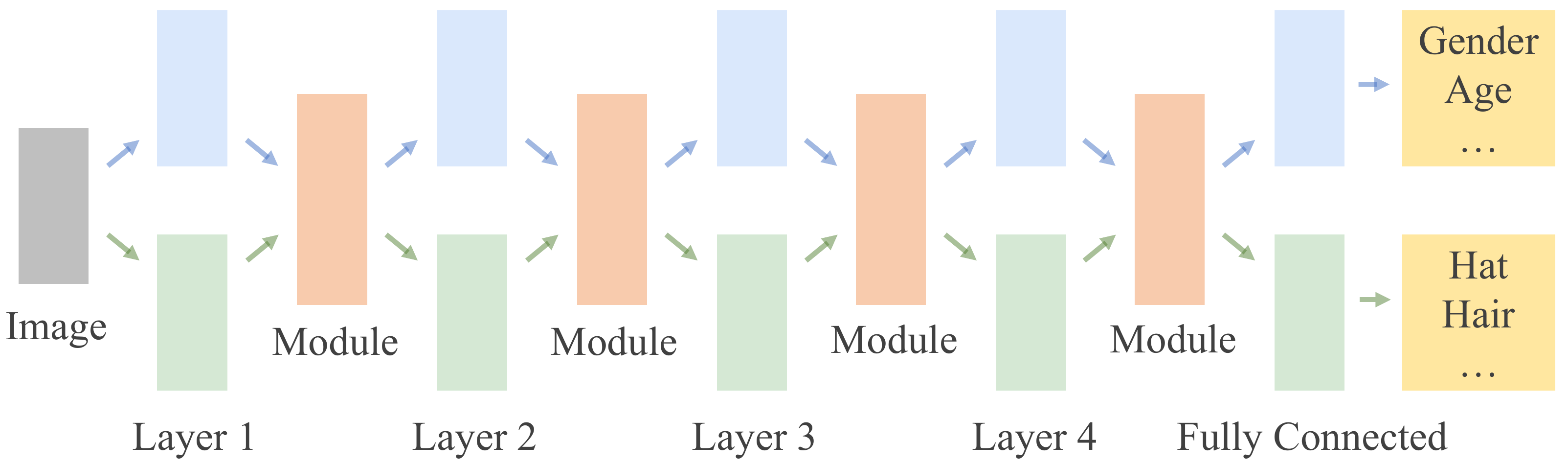}
        \label{fig:h_structure}
    }
    \caption{Different structures for multi-attribute recognition.}
    \label{fig:structures}
\end{figure}

Compared to the conventional image classification where an image 
only belongs to a single class, recognizing person attributes can be regarded as 
a multi-task (MT) learning problem, since a person is usually described with multiple attributes.
To recognize these attributes simultaneously, 
one popular approach is to build multiple attribute classifiers upon a shared backbone \cite{li2015multi}, as shown in Fig.~\ref{fig:shared_structure}, which is known as the hard parameter sharing structure \cite{ruder2017overview}.
However, this structure is prone to the negative transfer problem~\cite{pan2009survey, he2017adaptively, wang2019characterizing}, {\em i.e.}, 
the feature representation of one attribute may be impeded by dissimilar attributes if they are learned together.
To alleviate this problem, a vanilla approach is to ensemble two independent networks, as Fig.~\ref{fig:vanilla_structure}, and
each network predicts a more closely related subset of attributes \cite{hand2017attributes, cao2018partially}.
Nevertheless, in this structure there is no communication between two networks so that some useful correlations may be ruled out.
A more holistic paradigm is the soft parameter sharing structure \cite{ruder2017overview}, as illustrated in Fig.~\ref{fig:h_structure}, which absorbs advantages from both hard-sharing structure and vanilla structure.
It utilizes a certain module to decide what to share and what not to share with the other task at each layer.

\begin{figure*}
  \centering
  \includegraphics[height=64mm]{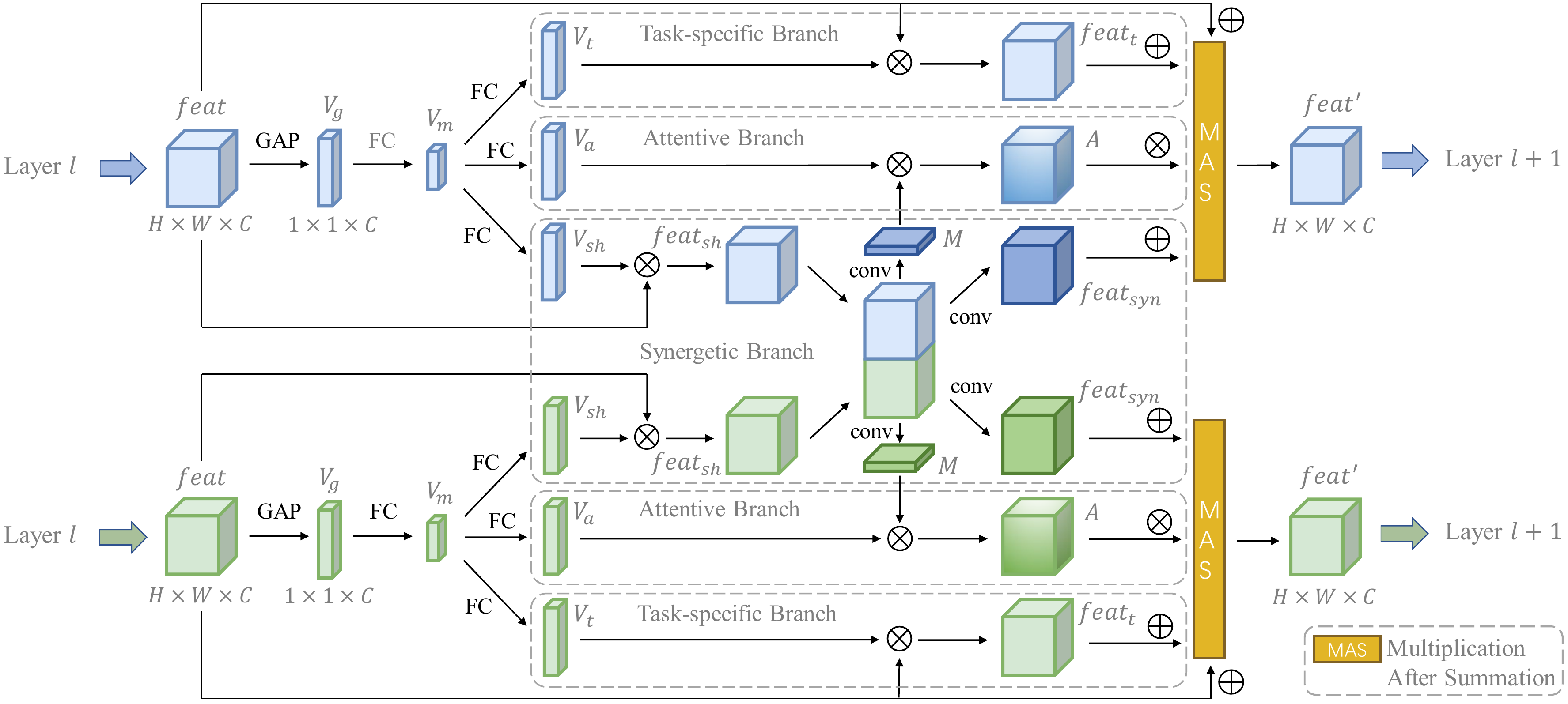}  
  \caption{Framework of the Co-Attentive Sharing module. It starts with two features $\mathrm{feat}$ from both networks in layer $l$. For each $\mathrm{feat}$, three channel attention $V_{sh},V_{a},V_{t} \in \mathbb{R}^{1 \times 1 \times C}$ are obtained with global average pooling (GAP) and fully connected (FC) layers, and are utilized in three branches.
  The synergetic branch produces enhanced features $feat_{syn}$ and spatial attention maps $M \in \mathbb{R}^{H \times W \times 1}$ using concantenated $feat_{sh}$ of two networks, while the attentive branch outputs a global attention $A$ and the task-specific branch outputs $feat_{t}$. Finally, outcomes of three branches are aggregated to get $feat^{'}$ as the input for layer $l+1$.}
  \label{fig:CAS}  
\end{figure*}

Apparently, sharing module is the most important component in soft-sharing structure.
Previous works like Cross Stitch \cite{misra2016cross} module and Sluice \cite{ruder122017sluice} module utilize linear interactions to enable feature sharing.
Cross Stitch module calculates the feature for next layer as a 
learnable linear combination of two input features.
Sluice module further divides the features into subspaces,
and obtains the new feature using a linear combination 
of subspaces.
However, in these methods, the interaction of features 
from different tasks is simply element-wise summation, so that the selection for discriminative channels is neglected.
Moreover, attributes usually correlate to different spatial locations of images \cite{li2018pose, liu2018localization}.
In other words, spatial information is non-trivial in recognizing pedestrian attributes.
However, the element-wise summation fails to utilize such information.
These two factors limit the performance of  feature sharing.

In order to better handle those challenges, 
we propose a novel \emph{Co-Attentive Sharing (CAS)} module which 
extracts discriminative channels and spatial regions for more 
effective feature sharing between two task networks 
in pedestrian attribute recognition.
It consists of three branches: \emph{synergetic} branch, 
\emph{attentive} branch and \emph{task-specific} branch. 
They exploit three different channel attentions generated from 
a shared intermediate vector and play different roles. 
The synergetic branch fuses the selected features from each 
task to generate enhanced features and spatial attention maps.
The attentive branch computes a global feature attention and 
the task-specific branch highlights important channels within each task.
Finally, the results of three branches is combined together as the
outputs of the module. 
Experiments on two large pedestrian attribute recognition datasets
demonstrate the effectiveness of the CAS module.






\section{Related Works}

\subsection{CNNs for Image Classification}
After the success of convolutional neural networks (CNNs) 
like AlexNet \cite{krizhevsky2012imagenet}, ResNet \cite{he2016deep}
on image classification and numerous other vision tasks, 
SENet \cite{hu2018squeeze} further improves the 
performance of CNNs by introducing a channel attention branch which 
models sophisticated channel inter-dependency. 
Moreover, BAM \cite{park2018bam} and CBAM \cite{woo2018cbam} propose 
novel methods to implement spatial attention in a convolutional module
integrated in the networks.
As recognizing attributes of a pedestrian is substantially a classification
problem, our work can benefit from their insights like emphasizing discriminative 
feature channels and spatial regions.

\subsection{Multi-task Learning for Facial Attribute Recognition}

In multi-task learning \cite{caruana1997multitask}, models which are trained for correlative tasks 
share complementary information with each other using certain mechanism.
Previous works like Cross Stitch \cite{misra2016cross} and Sluice \cite{ruder122017sluice} exploit automatically-learned linear combination to fuse features.
Multi-task learning gains popularity in facial attribute recognition 
since attributes are often semantically and spatially correlated.
Hand et al. \cite{hand2017attributes} propose a multi-task network which has shared bottom 
layers and bifurcates into branches for facial attribute classification. 
Cao et al. \cite{cao2018partially} introduce a Partially-Shared structure for multi-task learning,
where four task-specific networks exchanged complementary information with
each other through one shared network.
However, in most existing works
features from different tasks interact linearly so that the complex 
channel interdependency is less considered. 
Moreover, their sharing mechanisms are not capable to properly handle
spatial information.

\subsection{Pedestrian Attribute Recognition}

Early studies on pedestrian attribute recognition adopt
hand-craft features like color and texture channels \cite{deng2015learning}.
With the success of CNNs, recent works are usually based on deep models.
DeepMAR \cite{li2015multi} predicts several person attributes simultaneously with a shared backbone network.
Wang et al. \cite{wang2017attribute} propose an encoder-decoder architecture called JRL,
where one label is predicted given the context of all previously
predicted labels and attentive features.
HydraPlus-Net \cite{liu2017hydraplus} introduces a mechanism to select multi-scale 
and multi-semantic-level features for pedestrian attribute recognition.
PGDM \cite{li2018pose} uses an auxiliary pose estimation to better localize body-parts and enhance the performance.
In these works, the semantic correlation between attributes are well-modeled, and
the spatial information are also proved to be helpful for attribute recognition.
Nevertheless, a shared backbone is widely adopted and the influence of negative transfer \cite{pan2009survey, he2017adaptively, wang2019characterizing} has not received enough 
attention.


\section{Methodology}

In this section, we first introduce the Co-Attentive Sharing module, then describe the instantiated multi-task network.

\subsection{Co-Attentive Sharing Module}

Consider that two individual networks are trained for 
Task-A and Task-B respectively. The output features of layer $l$ of
each network, designated as
$feat^{A,l} \in \mathbb{R}^{H \times W \times C}$
and
$feat^{B,l} \in \mathbb{R}^{H \times W \times C}$,
are used as the inputs of the CAS module.
For each input, an intermediate vector $V_{m}$ is first obtained and serves as the basis for the following three branches, so that each branch can generate a channel attention vector from $V_{m}$.
Next, synergetic branch combines the selected features from both task
to obtain enhanced features $feat_{syn}$ and spatial attention maps $M$.
The attentive branch computes a global feature attention $A$ based on $M$,
and the task-specific branch highlights important channels within each task.
Finally, the results of three branches are aggregated as the input features
for layer $l+1$ of each task.
In following parts we introduce the details and omit the superscript 
of task and layer unless it is necessary for simplicity.

\textbf{Intermediate Vector.} 
First, the input convolutional feature $feat$ is first pooled into a
channel vector 
$V_{g} \in \mathbb{R}^{1 \times 1 \times C}$
using global average pooling (GAP). Then $V_{g}$ is fed 
into a linear (or fully connected, FC) layer $\mathbf{W}_{m}$ followed by a ReLU function to acquire 
the intermediate vector
$V_{m} \in \mathbb{R}^{1 \times 1 \times C/r}$
, where the number of channels is reduced with a ratio $r$.
Note that this process is just like the squeeze step in \cite{hu2018squeeze}, which produces a powerful channel descriptor for following branches.
In short, the intermediate vector is calculated as:
\begin{eqnarray}
  V_{m} = \mathrm{ReLU}(\mathbf{W}_{m}\mathrm{GAP}(feat))
\end{eqnarray}

\textbf{Synergetic Branch.} 
This branch aims to extract discriminative features and spatial attention
maps given the selected information from both tasks.
Towards this goal, a channel attention vector
$V_{sh} \in \mathbb{R}^{1 \times 1 \times C}$
is calculated by applying a linear layer $\mathbf{W}_{sh}$ and the sigmoid function on $V_{m}$, that is:
\begin{eqnarray}
  V_{sh} = \sigma(\mathbf{W}_{sh}V_{m})
\end{eqnarray}
where $\sigma$ is the sigmoid function. 
So the selected feature for cross-task sharing is computed as:
\begin{eqnarray}
  feat_{sh} = V_{sh} \otimes feat
\end{eqnarray}
where $\otimes$ stands for the element-wise multiplication. 
Note that the dimension of $V_{sh}$ is $1 \times 1 \times C$
and the dimension of $feat$ is $H \times W \times C$, during multiplication the attention vector $V_{sh}$ are
broadcasted (namely copied) along first two dimensions, and it is similar for other $\otimes$ operations.

Next, $feat_{sh}$ from two networks are concatenated 
along the channel dimension:
\begin{eqnarray}
  feat_{cat} = \mathrm{concat}(feat_{sh}^{A}, feat_{sh}^{B})
\end{eqnarray}
In order to fully utilize the information in $feat_{cat}$, 
two components are extracted from it for each network.
The first one is
$feat_{syn} \in \mathbb{R}^{H \times W \times C}$
, which contains discriminative 
feature representation from both networks.
It is calculated by applying a convolution layer with $1 \times 1$
kernel on $feat_{cat}$:
\begin{eqnarray}
  feat_{syn} = \mathrm{conv}^{1 \times 1}(feat_{cat})
\end{eqnarray}
The second one is a spatial attention map
$M \in \mathbb{R}^{H \times W \times 1}$. 
We follow Woo \emph{et al.} \cite{woo2018cbam} to produce this map:
\begin{eqnarray}
  M = \sigma(\mathrm{conv}^{7 \times 7}(\mathrm{concat}(\mathrm{Avg}(feat_{cat}),\mathrm{Max}(feat_{cat}))))
\end{eqnarray}
where $\mathrm{Avg}$, $\mathrm{Max}$ are mean and maximum functions
across the channel, $\mathrm{conv}^{7 \times 7}$ is a 
convolution with $7 \times 7$ kernel.

\textbf{Attentive Branch.}
Even though a spatial attention map is informative, intuitively it may be 
useful only for certain channels which need spatial regularization. 
Thus, in this branch, the spatial attention map $M$ from synergetic 
branch and a channel weight
$V_{a} \in \mathbb{R}^{1 \times 1 \times C}$ 
is used together to obtain a global attention
$A \in \mathbb{R}^{H \times W \times C}$. 
$V_{a}$ is another attention based on $V_{m}$, which is calculated by:
\begin{eqnarray}
  V_{a} = \sigma(\mathbf{W}_{a}V_{m})
\end{eqnarray}
where $\mathbf{W}_{a}$ is a linear layer.
And the global attention $A$ is:
\begin{eqnarray}
  A = V_{a} \otimes M
\end{eqnarray}

\textbf{Task-specific Branch.}
Besides the synergy branch where two networks exchange complementary
information,
this branch further improves the feature by strengthening the 
own feature of each task.
A vector is 
$V_{t} \in \mathbb{R}^{1 \times 1 \times C}$ 
is obtained similarly as before:
\begin{eqnarray}
  V_{t} = \sigma(\mathbf{W}_{t}V_{m})
\end{eqnarray}
where $\mathbf{W}_{t}$ is also a linear layer.
So that the outcome of the task-specific branch $feat_{t}$ is given by:
\begin{eqnarray}
  feat_{t} = V_{t} \otimes feat
\end{eqnarray}

\textbf{Final Aggregation.}
In order to get the final enhanced feature $feat^{'}$, 
we aggregate the results from each branch by first summing 
up $feat$, $feat_{syn}$, $feat_{t}$ and then multiply with $A$, 
that is:
\begin{eqnarray}
  feat^{'} = (feat \oplus feat_{syn} \oplus feat_{t}) \otimes A
\end{eqnarray}
where $\oplus$ denotes the element-wise addition.

\begin{figure}[b]
  \centering
  \includegraphics[width=82mm]{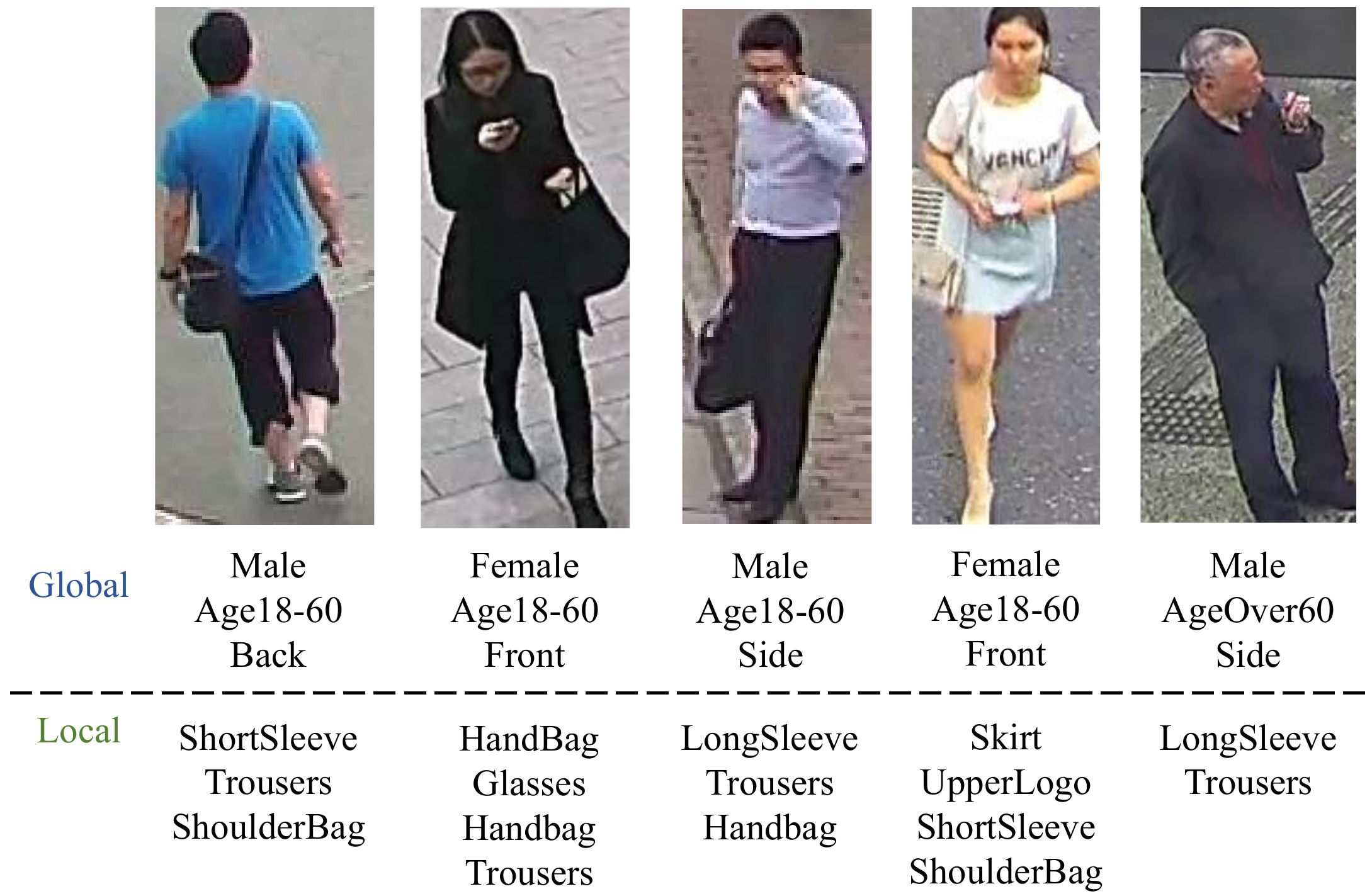}  
  \caption{Examples of the global-local grouping scheme.}
  \label{fig:default_group}  
\end{figure}

\subsection{Instantiation}

We instantiate the multi-task network using two pretrained ResNet-34 \cite{he2016deep}.
The CAS module is inserted on layer 1 to 4 for soft parameter sharing,
as depicted in Fig. \ref{fig:h_structure}.
The output of backbone network is fed into a linear
layer after global average pooling for predicting two groups of
attributes.

\section{Experiments}

\subsection{Experiment Setup}

\textbf{Datasets.}
We evaluate our method on two pedestrian attribute recognition datasets:
(1) PA-100K \cite{liu2017hydraplus} contains 100,000 pedestrian images with 26 annotated attributes
in total, which is the largest public pedestrian attribute recognition dataset as far as we know.
The dataset is split by 8$:$1$:$1 for training, validation and testing.
(2) PETA \cite{deng2014pedestrian} is a large-scale person attribute dataset with 19,000 images and 35 attributes.
There are 9,500, 1,900 and 7,600 images in training, validation and test set, respectively.

\textbf{Implementation.} We divide the attributes of both dataset into one \emph{global} group and one \emph{local} group for Task-A and Task-B respectively, shown in Fig. \ref{fig:default_group}.
Moreover, grouping scheme will not substantially
affect the performance, see Sec  \ref{sec:ablation_study}.
We adopt label-based metric $\mathrm{mA}$ and instance-based metric $\mathrm{Accuracy}$, 
$\mathrm{Precision}$, $\mathrm{Recall}$ and $\mathrm{F1}$ following \cite{li2015multi}.
The network is trained for 70 epochs with cross entropy 
loss and stochastic gradient descent. 
The learning rate is 0.02 and decays to 0.002 after 40 epoch.
Reduction rate $r$ is set to 16.

\subsection{Baselines and Competitors}

We setup four strong baselines for comparison:
(1) \textbf{Hard-Sharing:} a single ResNet-34 network which predicts all the attributes simultaneously.
(2) \textbf{MT-Vanilla:} the Vanilla structure with two independent networks that has no communication between each network.
(3) \textbf{MT-Cross-Stitch:} a soft-sharing structure 
with Cross Stitch module.
We integrate Cross Stitch module at the same places as our approach.
(4) \textbf{MT-Sluice:} a soft-sharing structure with Sluice module.
Our method is also compared to other pedestrian attribute recognition 
approaches \cite{wang2017attribute, liu2018localization, sarfraz2017deep, li2015multi, liu2017hydraplus, li2018pose, ji2019image}.

\begin{table}[t]
  \small
  \begin{center}
  \caption{Results on PA-100K.}
  \vspace{0.2cm}
  \begin{tabular}{p{80pt}|p{18pt}|p{18pt}|p{18pt}|p{18pt}|p{18pt}}
    \hline
    Method & mA & Acc. & Prec. & Recall & F1
    \\
    \hline
    \hline
    DeepMAR \cite{li2015multi} & 72.70 & 70.39 & 82.24 & 80.42 & 81.32\\
    HP-Net \cite{liu2017hydraplus} & 74.21 & 72.19 & 82.97 & 82.09 & 82.53\\
    VeSPA \cite{sarfraz2017deep} & 76.32 & 73.00 & 84.99 & 81.49 & 83.20\\
    PGDM \cite{li2018pose} & 74.95 & 73.08 & 84.36 & 82.24 & 83.29\\
    LG-Net \cite{liu2018localization} & \underline{76.96} & 75.55 & 86.99 & 83.17 & 85.04\\
    \hline
    Hard-Sharing & 75.43 & 76.53 & 87.85 & 83.38 & 85.56\\
    MT-Vanilla & 75.74 & 76.95 & 88.35 & 83.43 & 85.82\\
    MT-Cross-Stitch \cite{misra2016cross} & 76.55 & 77.30 & \underline{88.39} & 83.83 & 86.05\\
    MT-Sluice \cite{ruder122017sluice} & 76.26 & \underline{77.35} & 88.34 & \underline{84.10} & \underline{86.17}\\
    MT-CAS(ours) & \textbf{77.20} & \textbf{78.09} & \textbf{88.46} & \textbf{84.86} & \textbf{86.62}\\
    \hline
  \end{tabular}
  \label{tab:result_pa100k}
  \end{center}
\end{table}

\begin{table}[t]
  \small
  \begin{center}
  \caption{Results on PETA.}
  \vspace{0.2cm}
  \begin{tabular}{p{80pt}|p{18pt}|p{18pt}|p{18pt}|p{18pt}|p{18pt}}
    \hline
    Method & mA & Acc. & Prec. & Recall & F1
    \\
    \hline
    \hline
    DeepMAR \cite{li2015multi} & 82.89 & 75.07 & 83.68 & 83.14 & 83.41\\
    HP-Net \cite{liu2017hydraplus} & 81.77 & 76.13 & 84.92 & 83.24 & 84.07\\
    JRL \cite{wang2017attribute} & \textbf{85.67} & - & 86.03 & 85.34 & 85.42\\
    VeSPA \cite{sarfraz2017deep} & 83.45 & 77.73 & 86.18 & 84.81 & 85.49\\
    PGDM \cite{li2018pose} & 82.97 & 78.08 & 86.86 & 84.68 & 85.76\\
    IA$^{2}$-Net \cite{ji2019image} & \underline{84.13} & \underline{78.62} & 85.73 & \textbf{86.07} & \underline{85.88}\\
    \hline
    Hard-Sharing & 81.63 & 76.99 & \underline{86.89} & 83.56 & 85.20\\
    MT-Vanilla & 82.54 & 77.27 & 87.21 & 83.79 & 85.47\\
    MT-Cross-Stitch \cite{misra2016cross} & 82.02 & 77.66 & 87.17 & 84.40 & 85.76\\
    MT-Sluice \cite{ruder122017sluice} & 82.35 & 78.04 & 87.19 & 84.60 & 85.87\\
    MT-CAS(ours) & 83.17 & \textbf{78.78} & \textbf{87.49} & \underline{85.35} & \textbf{86.41}\\
    \hline
  \end{tabular}
  \label{tab:result_peta}
  \end{center}
\end{table}

\subsection{Experiment Results} 

\begin{figure}[t]
  \centering
  \includegraphics[height=47mm]{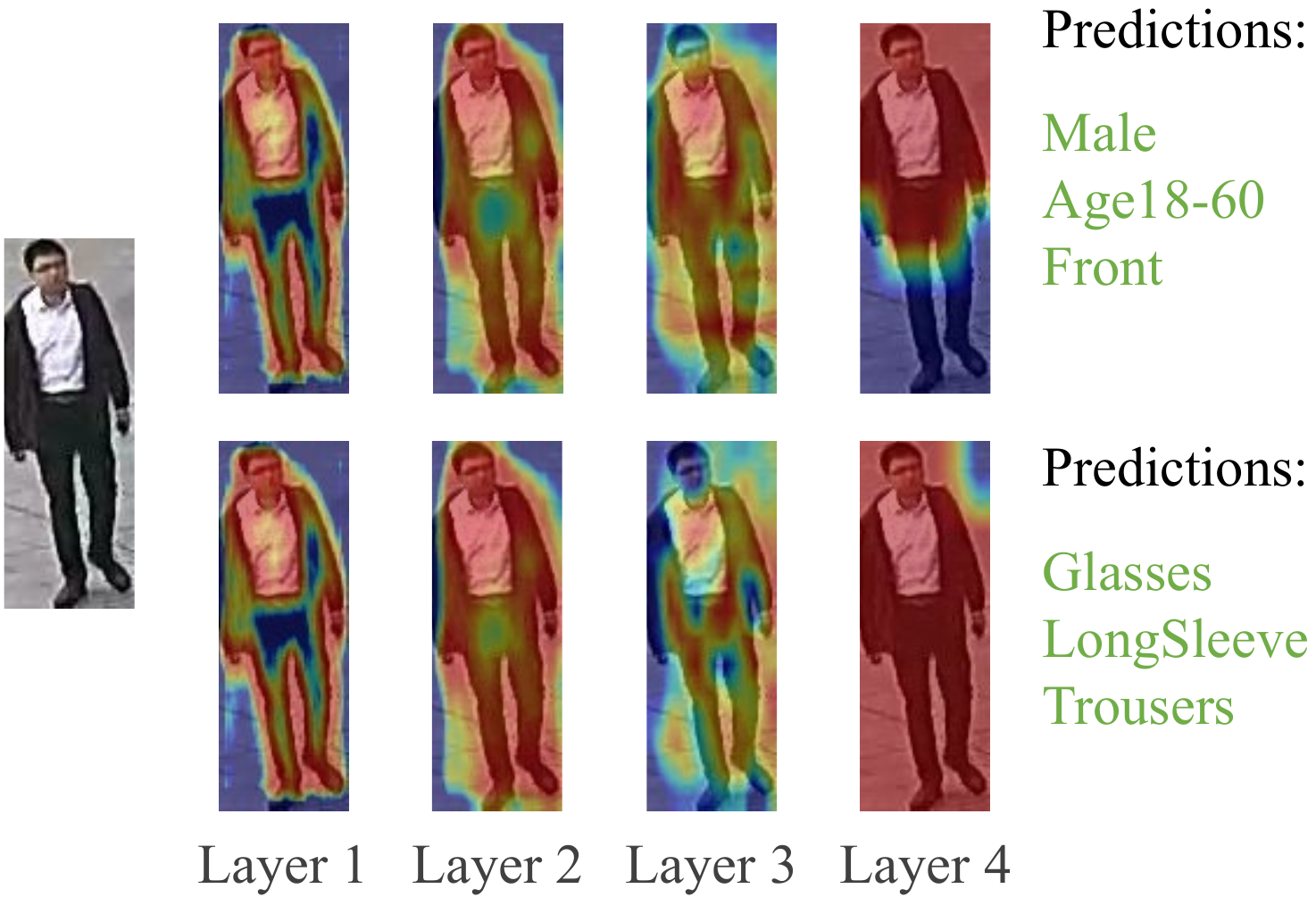}  
  \caption{Visualization of the spatial attention map $M$. The maps of two networks are similar in layer 1 and 2, and diverge in layer 3 and 4. In layer 3, Task A (upper) focus on the whole body, while Task B (below) emphasizes various smaller regions. In layer 4, the upper-body region is discriminative
  for the final prediction of Task A. All predictions are correct.}
  \label{fig:visualization_m}  
\end{figure}

The results are shown in Table \ref{tab:result_pa100k} 
and Table \ref{tab:result_peta}.
The highest score of each metric is marked in bold, and the second best one is underlined.
It is worth noting that our baselines have already surpassed a number of competitors.
We also provides qualitative results by visualizing the spatial
attention $M$ from each layer and each network, as shown in Fig. \ref{fig:visualization_m}.

\textbf{Comparison with Hard-Sharing.}
The MT-CAS model surpasses the Hard-Sharing 
baseline by 1.06\% on PA-100K and 1.21\% on PETA under $\mathrm{F1}$ metric, which demonstrates the effectiveness of the proposed framework.

\textbf{Comparison with MT-Vanilla.}
The MT-CAS model achieves a 0.80\% higher $\mathrm{F1}$ on PA-100K and 0.94\% higher $\mathrm{F1}$ on PETA, compared to MT-Vanilla. These results verify the effectiveness of CAS. Moreover, we also notice that MT-Vanilla slightly outperforms Hard-Sharing, however it does not fundamentally improve the results.

\textbf{Comparison with MT-Cross-Stitch and MT-Sluice.}
Our module also outperforms the Cross-Stitch and Sluice 
by a margin of about 0.5\% in $\mathrm{F1}$ score on PA-100K and about 0.6\% on PETA.
It shows that with the exploitation of channel and spatial attentions, our module is capable to share more discriminative features than previous methods.

\subsection{Ablation Study} \label{sec:ablation_study}

To understand the effectiveness of each key component of CAS module and the influence of other factors, we conduct further analysis using PA-100K dataset.

\textbf{Synergetic Branch.}
We replace the original feature exchange operation with two alternates.
One is to use element-wise summation instead of concatenation to 
aggregate $feat_{sh}^{A}$ and $feat_{sh}^{B}$, that is 
$feat_{cat}^{-} = feat_{sh}^{A} \oplus feat_{sh}^{B}$.
The other one is to further remove the $\mathrm{conv}^{1 \times 1}$ which leads to
$feat_{syn}^{A} = feat_{syn}^{B} = feat_{cat}^{-}$.
We refer to them as $\mathrm{Synergetic}^{-}$ and 
$\mathrm{Synergetic}^{--}$.
Results in Table \ref{tab:branch} reveal that
concatenating and convolution operation are important for improvement.

\textbf{Attentive Branch.}
We remove the $V_{a}$ and use $M$ as the global attention through
broadcasting in final aggregation.
In another case, we remove the entire attention branch from the module.
These two modifications denotes $\mathrm{Attentive}^{-}$ and
$\mathrm{Attentive}^{--}$, respectively.
In Tabel \ref{tab:branch}, we observe that without attentive branch, the performance drops by about 0.3\%. We also notice that the spatial attention does not improve the performance if it is not used with channel attention. It indicates that channel attention is critical in spatial regularization.

\textbf{Task-specific Branch.}
We ablate the whole Task-specific branch, which is 
designated as $\mathrm{TS}^{--}$. The influence of removing 
this branch is about 0.2\% on $\mathrm{F1}$, shown in Table \ref{tab:branch}.

\begin{table}[b]
  \small
  \begin{center}
  \caption{Ablation study results of key components.}
  \begin{tabular}{p{73pt}|p{20pt}|p{20pt}|p{20pt}|p{20pt}|p{20pt}}
    \hline
    Method & mA & Acc. & Prec. & Recall & F1
    \\
    \hline
    \hline
    Synergetic$^{-}$ & 76.77 & 77.70 & 88.24 & 84.53 & 86.35\\
    Synergetic$^{--}$ & 76.53 & 77.64 & 88.26 & 84.38 & 86.27\\
    \hline
    Attentive$^{-}$ & 76.90 & 77.62 & 88.33 & 84.40 & 86.32\\
    Attentive$^{--}$ & 77.02 & 77.67 & 88.25 & 84.51 & 86.34\\
    \hline
    TS$^{--}$ & 77.18 & 77.87 & 88.27 & 84.63 & 86.41\\
    \hline
    Channel$^{--}$ & 76.99 & 77.61 & 88.02 & 84.52 & 86.23\\
    \hline
    Full Module & 77.20 & 78.09 & 88.46 & 84.86 & 86.62\\
    \hline
  \end{tabular}
  \label{tab:branch}
  \end{center}
\end{table}

\textbf{Channel Attention.}
We remove all channel attentions $V_{sh}$, $V_{t}$, $V_{a}$ so that $feat_{sh}^{-}=feat$, and use broadcasted $M$ as $A$. This case denotes $\mathrm{Channel}^{--}$ in Table \ref{tab:branch}, which demonstrates the effectiveness of selecting discriminative channels.

\textbf{Grouping Scheme.}
We setup three additional schemes by different principles mentioned in \cite{wang2017attribute}: \emph{rare-frequent}, \emph{top-down} and \emph{random}. In first two well-designed schemes, attributes in each group are more relevant, while in the random scheme dissimilar attributes are more likely to be put into one group.
Results of first two schemes are close to original MT-CAS
while the random one has a decline of 0.3\% on $\mathrm{F1}$ score, indicating that a proper grouping scheme is helpful.

\begin{table}[h]
  \small
  \begin{center}
  \caption{Influence of grouping scheme.}
  \vspace{0.2cm}
  \begin{tabular}{p{70pt}|p{20pt}|p{20pt}|p{20pt}|p{20pt}|p{20pt}}
    \hline
    Grouping Scheme & mA & Acc. & Prec. & Recall & F1
    \\
    \hline
    \hline
    rare-frequent & \textbf{77.34} & \textbf{78.07} & \textbf{88.61} & \underline{84.64} & \textbf{86.58}\\
    top-down & 76.96 & \underline{77.96} & \underline{88.40} & \textbf{84.69} & \underline{86.51}\\
    random & \underline{77.01} & 77.71 & 88.26 & 84.53 & 86.35\\
    \hline
  \end{tabular}
  \label{tab:group_scheme}
  \end{center}
\end{table}

\textbf{Reduction rate.}
Reduction rate $r$ controls the dimension reduction of $V_{m}$
with respect to the original channel number $C$. 
We try $r$ from 2 to 32.
The results in Table \ref{tab:reduction_rate} show that generally the most appropriate choice is $r=16$.

\begin{table}[h]
  \small
  \begin{center}
  \caption{Influence of reduction rate.}
  \vspace{0.2cm}
  \begin{tabular}{p{70pt}|p{20pt}|p{20pt}|p{20pt}|p{20pt}|p{20pt}}
    \hline
    Reduction rate & mA & Acc. & Prec. & Recall & F1
    \\
    \hline
    \hline
    $r=2$ & \textbf{77.39} & \underline{78.02} & \textbf{88.55} & 84.64 & \underline{86.55}\\
    $r=4$ & \underline{77.24} & 77.71 & 88.40 & 84.33 & 86.32\\
    $r=8$ & 77.11 & 77.92 & \underline{88.46} & 84.63 & 86.50\\
    $r=16$ & 77.20 & \textbf{78.09} & \underline{88.46} & \textbf{84.86} & \textbf{86.62}\\
    $r=32$ & 77.16 & 77.82 & 88.18 & \underline{84.70} & 86.41\\
    \hline
  \end{tabular}
  \label{tab:reduction_rate}
  \end{center}
\end{table}

\textbf{Module integration.}
We modify the integration positions of the module.
As the number of combinations is not small, we experiment
with cases where modules are placed consecutively.
From Table \ref{tab:integration}, we see that the most effective positions for integration are layer 2 and 3.
Generally, the more CAS modules are inserted in a network, the higher $\mathrm{F1}$ it achieves.

\begin{table}[h]
  \small
  \begin{center}
  \caption{Influence of module integration positions.} 
  \vspace{0.2cm}
  \begin{tabular}{p{30pt}p{30pt}p{30pt}p{30pt}p{40pt}}
    \hline
    \makecell[c]{Layer 1} & \makecell[c]{Layer 2} & \makecell[c]{Layer 3} & \makecell[c]{Layer 4} & \makecell[c]{F1}
    \\
    \hline
    \hline
    \makecell[c]{$\checkmark$} & & & & \makecell[c]{86.14} \\
    & \makecell[c]{$\checkmark$} & & & \makecell[c]{\textbf{86.36}} \\
    & & \makecell[c]{$\checkmark$} & & \makecell[c]{\underline{86.29}} \\
    & & & \makecell[c]{$\checkmark$} & \makecell[c]{86.11} \\
    \hline
    \makecell[c]{$\checkmark$} & \makecell[c]{$\checkmark$} &  &  & \makecell[c]{\underline{86.39}} \\
    & \makecell[c]{$\checkmark$} & \makecell[c]{$\checkmark$}  &  & \makecell[c]{\textbf{86.50}} \\
    & & \makecell[c]{$\checkmark$} & \makecell[c]{$\checkmark$}  & \makecell[c]{86.37} \\
    \hline
    \makecell[c]{$\checkmark$} & \makecell[c]{$\checkmark$} & \makecell[c]{$\checkmark$} & & \makecell[c]{\textbf{86.54}} \\
    & \makecell[c]{$\checkmark$} & \makecell[c]{$\checkmark$} & \makecell[c]{$\checkmark$} & \makecell[c]{86.50} \\
    \hline
    \makecell[c]{$\checkmark$} & \makecell[c]{$\checkmark$} & \makecell[c]{$\checkmark$} & \makecell[c]{$\checkmark$} & \makecell[c]{\textbf{86.62}} \\
    \hline
  \end{tabular}
  \label{tab:integration}
  \end{center}
\end{table}

\section{Conclusion}
In this paper, we introduce the CAS module which enables soft feature sharing between two networks 
for person attribute recognition.
The CAS module consists of three branches of different functions. It leverages channel and spatial information in two-task feature sharing, which is less-considered in previous works.
The experimental results on two large pedestrian attribute recognition datasets show that the module outperforms the hard-sharing structure and two representative soft-sharing structures.
Furthermore, extensive studies verify the effectiveness of each key component of the module.

\section{Acknowledgement}
This work was supported by the Natural Science Foundation of China (Project Number 61521002 and 60673107).

\small
\bibliographystyle{IEEEbib}
\bibliography{references}

\end{document}